\documentclass{article}
\usepackage{arxiv}
\usepackage{graphicx}

\begin{document}

\title{ Mind Reading at Work: Cooperation without Common Ground }

\author{Peter Wallis}



\maketitle

\begin{abstract}
As Stefan Kopp and Nicole Kr\"amer say in their recent
paper~\cite{KoppKramer21}, despite some very impressive demonstrations over the
last decade or so, we still don't know how how to make a computer have a half
decent conversation with a human.  They argue that the capabilities required to
do this include incremental joint co-construction and mentalizing.  Although
agreeing whole heartedly with their statement of the problem, this paper argues
for a different approach to the solution based on the ``new'' AI of situated
action.
\end{abstract}

\section{Introduction}
One problem with developing human-computer interfaces is that it involves
both engineers and humans. Human behaviour is a tough nut to crack for
engineers and those who are expert at human behaviour are rarely able to put
things in engineering terms.  Kopp and Kr\"amer's paper gives a succinct
introduction to why we need more theory and following Tomasello~\cite{tomasello08}, they say that we need to pay attention to the
{\itshape intentional} and {\itshape cooperative} nature of Human-Human
Interaction (HHI). However saying that language is intentional and cooperative
is simply not as quantitative as scores on a leader board.
The aim of this paper is to first translate the problem of human-agent
interaction (HAI) and conversational user interfaces into terms engineers will
understand, and second, to point out a significant development by engineers and
how it applies to this grand challenge.
In particular, whereas many readers of the Kopp and Kr\"amer paper will assume
``the common ground'' must be represented symbolically, engineers are at least
sceptical about the notion of symbolic representation and these days have
a much better idea of how to make computers reason about action.

\section{What we know: the problem}

From an engineering perspective, first, we know that communication is 
{\itshape always} with respect to a model.  From Information Theory we know,
for instance, that you can communicate the text of ``War and Peace'' with
a single bit of data, if you know that the message is either War and Peace or
something else -  The Collected Works of Shakespeare perhaps.
Compression and encryption techniques rely on having the sender
and receiver use {\itshape exactly} the same model.  In the early days of NLP
it was thought that Russian was simply an encrypted version of English and all
that was needed for machine translation was to break the code.  The underlying
assumption was that all humans share the same model.  And as a successful
human it seems obvious that, if I order a peperoni pizza with sweet peppers and
no olives that, like assembling Lego, the receiver of my order will be able to
ground ``sweet peppers'' to something in the world that will end up on my pizza.

It turns out that HHI does not work that way.  The initial assumption in NLP
was that humans do indeed share models with enough detail to decode language,
but that the language itself was imprecise.  The way forward was to map sloppy
natural languages into something better that, presumably, looked like
predicate calculus.  The notion of primitive acts~\cite{sch73} suggested that
words like ``sell'' could be mapped into a longer string of more concise
``logical atoms'' such as A p-trans B to C; C m-trans D to A; where B is the
goods and D is the money.  The realisation was that there is an awful lot of
this kind of knowledge~\cite{all87}.  There was however plenty of interest in
doing the work. Wilks convinced Longmans to provide their Dictionary of
Contemporary English in a form that used a limited defining vocabulary (on
magnetic tape!) that might go some way toward identifying a set of semantic
primitives~\cite{wil89}. It also led to using statistical models of corpora
- techniques that today would be marketed as Machine Learning - and to
ambitious projects such as the CYC project~\cite{cyc} which aimed to hand code
the required facts.

Looking back we can say that this programme failed. The failure was however 
a significant contribution by AI research in that it goes against a long
standing tradition that natural languages are some kind of fallen version of
something better. An idea that goes back to biblical times and was the
motivation for formal logic.  More significantly perhaps we now know - or
should know - what not to try.

The most recent band of interest in conversational user interfaces puts the
focus, not on understanding the problem, but on engineering a solution.
Neural Nets and indeed statistical models of dialog avoid the theory by
talking of deep learning, hidden layers~\cite{pomdp07} with only
speculation to back up any explanation of how it works (when it does work).
This is fine as far as it goes, but there is also a tendency, especially under
the pressure of conditional funding, to tailor the problem to fit the solution.
Looking at the text of the latest ``gold standard'' dialog system evaluation
data~\cite{budzianowski2020multiwoz}, the tasks are heavily biased toward
``understanding'' the input text where understanding is defined to be just slot
filling. This is in stark contrast to the DARPA Communicator
project~\cite{CommEval} and the Alexa Prize~\cite{AlexaPrize} which were
serious attempts to advance the state of the art -- as indeed was the initial
Loebner Prize~\cite{shie94}.
Rather than some made up metric, the Amazon team - with plenty of experience,
motivation, and resources - have opted for an evaluation based on human
judgement.  Human judgement being of course the ultimate, theory free, arbiter
of success.  While the ML people play with their data, the theory behind
language in use does seem to be on hold.  And as the limited progress on the
Alexa Prize and its predecessors suggests, it would seem theory is what is
needed.

As Kopp and Kr\"amer suggest, the current understanding is that we need a
better way for our machines to assemble the common ground in a conversation.
 Without a shared model to work from, an Information Theoretic perspective
suggests HHI builds a shared model as the conversation progresses. The common
ground is built on-the-fly.

\section{Where we had got to}
Those who study HHI have made some observations that may help.  First, a human
is very good at recognising the {\itshape intent} of his or her CP. That is, we
are very good at guessing what our CP wants to do.  Note the use of ``intent''
here is the popular one as in ``I intend to pick up some milk on the way home''
and {\itshape not} the referential ``intent'' used in the broader research
community as a way of avoiding talk about meaning of words.
Note also that recognition of intention is not a skill unique to humans - nor
even the great apes. Children develop a ToM - a Theory of Mind in others - from
about the age of three, and a sheep dog is very good at guessing where sheep
are about to go.
Our ability to recognise the intentions of others is not perfect by any means
but, second, humans are {\itshape compelled} to work hard to sort out communication
failures.  That is, they cooperate in the process of understanding, even if
they disagree about the content of that communication. 
Consider this (naturally occurring) example from Eggins and Slade talking
about sequential relevance:
\begin{quote}
	\begin{tabular}{r|p{60mm}}
	\textbf{A:} & What's that floating in the wine?\\
	\textbf{B:} & There aren't any other solutions.\\
	\end{tabular}

	\noindent
	You will try very hard to find a way of interpreting B's turn as somehow 
	an answer to A's question, even though there is no obvious link between
	them, apart from their appearance in sequence. Perhaps you will have
	decided that B took a common solution to a resistant wine cork and
	poked it through into the bottle, and it was floating in the wine.
	Whatever explanation you came up with, it is unlikely that you looked
	at the example and simply said `it doesn't make sense', so strong is
	the implication that adjacent turns relate to each
	other~\cite{EggSla97}.
\end{quote}
Tomasello's point is that the great apes would look at the utterance, say it
didn't make sense, and move on.  Humans on the other hand, are compelled to
try and figure it out.  This human behaviour is normative. As Tomasello points
out~\cite{tomasello08}, if an individual in society fails to understand the
utterances of others they will be considered simple and lock up. If he or she
produces utterances that others cannot understand, then they will be locked up
as mad.  In order to be treated as human, one {\itshape must} be able to ``read
minds'' and actively cooperate in the communicative processes of the
people around you.

The classic approach at this point would be to implement a Good Old Fashioned
AI (GOFAI) model of ToM, literally modeling the mind of the system's
conversational partner. This approach was developed in James Allen's TRAINS
programme and turns out to be an extremely costly.  There is an alternative
approach to ToM that is discussed below but before getting to that however,
it is useful to take a step back and look at how HHI works and why something
like a ToM is critical.

\section{How language works}

The above provides an introduction to the problem but engineers tend to want
details.  In a rare case of an algorithm from the human sciences,
Seedhouse~\cite{Seedhouse04} summarises the findings of Conversation Analysis
as follows. Once a conversation is established between a Speaker and his or her
Conversational Partner, at any point:
\begin{itemize}
\item
the speaker's utterance will go {\bfseries seen but unnoticed} if the
utterance is the second part of an adjacency pair - an actual answer to
a question or a greeting in response to a greeting. This is the type of
interaction we know how to do with a machine.  If I walk in to
a corner shop and ask for ``six first class stamps please'', the person behind
a counter in the UK is no doubt going to know exactly what I mean, and so can
a machine.
\item
Often/sometimes however, a speaker's utterance will not ``fit'' with what went
before.  Taking a classic example from the CA literature, consider someone
walking into a corner shop and asking ``Excuse me. Do you sell stamps?'' and
the person behind the counter says ``first or second class?''.  The shop
keeper's response is not an answer to the question.  Instead the shop keeper
moved on, and the customer automatically {\bfseries noticed and account-for}
the speaker's utterance - and account-for it effortlessly.  As described above,
we humans are compelled to cooperate.  The shop keeper is compelled to assume
that what was heard made sense to the speaker, and ``works hard'' to figure it
out.
\item
If the conversational partner cannot account for the speaker's
utterance, then the speaker {\bfseries risks sanction}.  If, in a breaching
experiment, I walk into a shop and ask for ``one sixth class stamp please'', the
person behind the counter does not say ``Sorry I do not have any of those,''
but instead might say something like ``Sorry?'', meaning please repeat that.
What happens next is complex, but not intractable.  For instance I might
say ``Six first class stamps'' and the person behind the counter might
{\itshape account for} my odd behaviour by deciding I had made an unconscious
spoonerism.  The process of bringing the conversation back on track is automatic
and highly dependent on things like social distance and power relations.
Things that we all know how to do even if we don't notice we're doing it.
\end{itemize}

We know how to make computers do the seen but unnoticed, and we are all
socialized humans who know how to ramp up toward sanction when needed.  The
hard part is making machines that can account-for a speaker's utterance that
succeeds and fails as humans expect it to.  It is here that mind reading
becomes critical.

The purpose of this paper is to present an alternative to the idea that we need
to implement some model of Theory-of-Mind in order to do the accounting-for.
Key to this alternative is that ``mind reading'' happens within this
seen-but-unnoticed / noticed-and-accounted-for / works-toward-sanction
framework. For want of a better acronym, let us call this the sbu/naaf/wts
loop.

\section{Mental Models and Accounting for}

In the late 1980's through the 1990's Brooks' robots~\cite{brooks91} caused a
revolution in AI when he abandoned planning in favour of a reactive
architecture that became Behaviour-Based Robotics~\cite{Ark98}.  Rather
than sensing the world, modeling it, then planning and acting,  a software
unit here called a ``behaviour'' connects sensing to action in a continuous
process. Any reasoning - often also quite reactive - then inhibits, stimulates, or in other ways modifies, the parameters of the behaviour.  This, it turns
out, is how real situated agents work, anything from the bug spotting neurons in frogs, to humans using a photocopier~\cite{suchman87}, through to trans
national corporations managing their supply chain~\cite{HeJa96}. 
The proposal is that rather than reasoning about other minds, we should figure
out a situated approach to mind reading.

The enactivists have indeed been doing this and talk about ``interpersonal
primary subjectivity,'' writing books to explain what this is and how it
addresses various problems~\cite{Gallagher20}. The gist, as I see it, is that
when I see a helicopter hover over head, I actually see at most half a
helicopter, but if you ask me I would say I perceive a helicopter. Perception
comes with assumptions and hence may be wrong, but the process is automatic.
A hovering metal orb may turn out to be a soup ladle~\cite{orb}, but I
have already {\itshape perceived} a hovering orb in the same way as I directly
perceive the helicopter.
The enactivist argument is that we humans {\itshape directly perceive} the
intent of others.  What we {\itshape see} is things like where their attention
is focused, their prior actions that led to their current potential for action,
and so on, but the process of turning that into a perception is, like walking
on two legs, hardwired.. or at least in the firmware.

This is all well and good, but like Brooks' robots, the philosophy and
psychology tend to support the idea that evolution works on the engineers' KISS principle\footnote{KISS -- keep it simple stupid.}.

\section{An engineering approach}

Our epiphany was to realise that for task based dialogs there is no point
recognising the intent of the system's conversational partner if the system
does not have a means of dealing with it~\cite{WalEdm17}.  This is especially
true given intention recognition happens within the sbu/naaf/wts loop, giving
our agent a (socially proscribed) means of signaling if the mechanism fails.
Rather than planning from first principles as the TRAINS system did, we
proposed a dialog management system that works from a library of behaviours,
each of which addresses a use-case. The sleight-of-hand here is that use-cases
are normally associated with some goal, and by tagging the behaviour with the
relevant goal, we can fake mind reading.

Consider a system that enables me to ring up and order a pizza.  If I order
a peperoni pizza with sweet peppers and no olives, and the system knows how to
create such a pizza, then the process is straight forward (the seen but
unnoticed).  If I ring the service and order a taxi - something for which our
pizza ordering system cannot, quite reasonably, account for - the system can
work toward sanction. It might, for instance say ``you have called Papa Gino's
Pizzeria. How can I help?'' which as a speech act is, among other things,
a refusal to talk about taxis (see \cite{wallis15} for a detailed discussion).
Recognising what the caller wants when it is reasonable, but not explicit, is
the hard part.

So imagine the situation where the caller's opening statement is "Peperoni,
peppers, no olives". We humans expect the caller wants to order a pizza even
though he has not said that.  The standard HCI approach would be to design in
the assumption and have the system present itself as all and only a pizza
ordering system. This way it is the user's fault if he tries to do something
else with it.  An alternative is to acknowledge that the user is more than
a user -- users are also humans, embedded in a soup of social relations that
need to be negotiated.  The system is thus expected to negotiate a shared
intent, which requires some level of mind reading.

The system we proposed has a library of behaviours, each tagged with the goal
it achieves.
\begin{figure}[h]
  \centering
  \includegraphics[width= 0.8\textwidth]{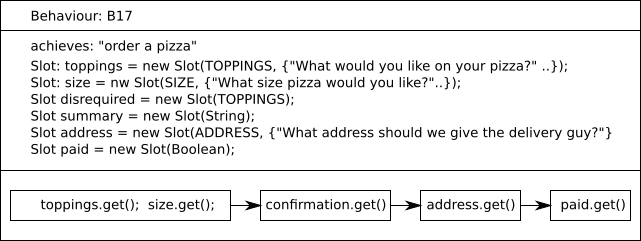}
  \caption{A behaviour, task~\cite{gray2019craftassist}, plan~\cite{MaxGuide},
  or activity~\cite{mix04}, for ordering a pizza. Note the conversational
  opening and closing are handled by the separate sbu/naaf/wts loop.}
\end{figure}
Let's say that of those behaviours, B17 has, among other things, 
slots for required and disrequired toppings, and that match pepperoni and
peppers, and olives. In addition B17 (see Figure) is tagged with the goal "order a pizza".
If B17 is identified as the behaviour with the best match with the user's
utterance above, the system can respond with:
``You want to \textless goal\textgreater order a pizza\textless
/goal\textgreater. Great! \textless description\textgreater pepperoni ...''.
The system, with no more than classic slot filling techniques and
a use-case, has performed what looks like mind reading.

 This is no doubt completely underwhelming to many and that is exactly the
point. For us humans the process is automatic, but consider how the same result
would have to be achieved with a GOFAI approach.  There is no description of
a pizza in the opening statement and only a bunch of ingredients in the common
ground.  With the GOFAI approach, the user and the system co-construct a shared
model of a pizza, and then the system has a separate planning system to decide
how to make the thing represented.  Here there is no shared model, but rather
two models of what is being talked about.  The caller's head {\itshape might}
contain a model of a pizza, but for the system the pizza is represented by
specifications for a process.

\subsection{A minor extension}
The argument is that this model of mind reading is not very robust but that is
not critical because the whole process operates within the sbu/naaf/wts loop.
If the system gets it wrong, the human will let the machine know by working
towards sanction.  The challenge is to recognise this. The important point is
that the human is {\itshape compelled} to do this in HHI, and will reliably
continue to do this as long as he or she is treating the agent as a human in
the interaction.

Consider the case where I ring up and say I would like to order a peperoni
pizza with sweet peppers and absolutely no olives. In this case the
``absolutely'' may seem rather redundant.  The perfect system - the theory
about how HAI ought to work - would actively account-for everything in the
user's utterance that does not go seen-but-unnoticed.  In the terminology of
Conversation Analysis, the system would actively seek out an answer the
question ``why this, in this way, right here, right now?'' The
appearance of ``absolutely'' is anomalous and so why is it there? Like the wine
bottle example above, the system may simply decide it doesn't make sense and
move on, but if our pizza ordering system has a behaviour for checking for
allergies, this might be a good time for the system to use it. The mechanism
for triggering this could be simply key-word spotting, but a post-hoc
explanation would be based on the goal associated with the allergy checking
behaviour. A follow-up question by the system might be ``No olives. Olive oil
is okay?'' and the caller might say ``sure'' in a way that suggests that
is an odd question.  The system might then help the caller account-for its
utterance by saying "I just wanted to \textless goal\textgreater check for
allergies\textless/goal\textgreater''.
Although the evidence for an allergy is weak, the interactive nature of dialog
means that getting it wrong here does not matter. As long as the caller can
account-for the system's utterances, the system does not bring on sanction.

In combination with the sbu/naaf/wts loop, our engineer's solution to mind
reading is more than a cheap trick and indeed provides an enactivist model of HHI.

\section{Summary}
Making a machine capable of intention recognition is difficult using the
techniques of classic AI, but we now know that classic AI was wrong about how
minds work.  Rather than conversational partners constructing a shared model of
a pizza on common ground, a conversation is the negotiation of independent
models of what the relevant CP {\itshape wants}.  Rather than grounding
language in apparently objective {\itshape things}, the above pizza making
agent grounds language in the {\itshape doing}, as represented by use cases.
The way people think about doing, is in terms of goals, and goals are thus
{\itshape inherent} in the way we use language to communicate.  Recognising
the intentions of others is not a reliable process, but the normative structure
of human communication as represented here by the sbu/naaf/wts framework
provides a robust repair mechanism.  This approach is not only practical for
HAI but also based on current trends in HHI.

  \bibliographystyle{acm}

\end{document}